# ErgoChat – a Visual Query System for the Ergonomic Risk Assessment of Construction Workers

Chao Fan[1], Qipei Mei[2], Xiaonan Wang[3], and Xinming Li[1, *]

[1] Department of Mechanical Engineering, University of Alberta, 116 St NW, Edmonton, Alberta T6G 2E1, Canada

[2] Department of Civil and Environmental Engineering, University of Alberta, 116 St NW, Edmonton, Alberta T6G 2E1, Canada

[3] Department of Electrical Engineering, University of Alberta, 116 St NW, Edmonton, Alberta T6G 2E1, Canada

*Corresponding author's email address: xinming.li@ualberta.ca

**Abstract**

In the construction sector, workers often endure prolonged periods of high-intensity physical work and prolonged use of tools, resulting in injuries and illnesses primarily linked to postural ergonomic risks, a longstanding predominant health concern. To mitigate these risks, researchers have applied various technological methods to identify the ergonomic risks that construction workers face. However, traditional ergonomic risk assessment (ERA) techniques do not offer interactive feedback. The rapidly developing vision-language models (VLMs), capable of generating textual descriptions or answering questions about ergonomic risks based on image inputs, have not yet received widespread attention. This research introduces an interactive visual query system tailored to assess the postural ergonomic risks of construction workers. The system's capabilities include visual question answering (VQA), which responds to visual queries regarding workers' exposure to postural ergonomic risks, and image captioning (IC), which generates textual descriptions of these risks from images. Additionally, this study proposes a dataset designed for training and testing such methodologies. Systematic testing indicates that the VQA functionality delivers an accuracy of 96.5%. Moreover, evaluations using nine metrics for IC and assessments from human experts indicate that the proposed approach surpasses the performance of a method using the same architecture trained solely on generic datasets. This study sets a new direction for future developments in interactive ERA using generative artificial intelligence (AI) technologies.

## 1 Introduction

Prompt and effective identification and mitigation of workplace hazards are essential for maintaining safety, health, and productivity within the work environment. In the construction industry, workers are often subject to conditions that require awkward body postures, repetitive motions, and intense physical effort, which can detrimentally impact their health [1]. Such conditions in construction tasks usually lead to the emergence of work-related musculoskeletal disorders (WMSDs). Statistics from the United States Bureau of Labor Statistics show that the construction industry's injuries and illnesses caused by WMSDs ranked fifth among all industries. Moreover, in the same year, WMSDs represented 30% of all occupational injuries and illnesses [1]. According to the Association of Workers' Compensation Boards of Canada, the manufacturing and construction sectors reported the second and third-highest rates of lost-time injury claims in 2021, representing 13.6% and 10.4% of claims, respectively [2]. European Agency for Safety and Health at Work indicated that the construction and manufacturing sectors reported the highest sick leave rates due to WMSDs [3]. Therefore, early identification and proactive prevention of WMSDs caused by postural ergonomic risks are crucial for companies and construction industry workers.

Traditional methods of postural ERA include self-reporting, observation, and direct measurements [4–11]. Self-reporting and observational methods are limited by the need for manual data collection, which includes the significant time required by ergonomic experts and inconsistencies in data interpretation. Alternatively, the direct measurement method employs sensors or markers attached to workers and uses a motion capture system to evaluate their postural ergonomic risks. While this type of method is more time-efficient and objective compared to self-reporting and observational techniques, the attached sensors or markers can create psychological and physical burdens on the workers, potentially impacting their work performance. Additionally, motion capture systems that use markers demand specific environmental conditions to function effectively, such as adequate ambient lighting and clear visibility of markers. In summary, the limitations inherent in traditional approaches substantially detract from the effectiveness of ERA methods.

To overcome the challenges of subjectivity, inconsistency, intrusiveness, and the time-intensive nature of traditional ERA methods, predictive AI has become a popular tool for researchers. Predictive AI encompasses AI systems that employ statistical analysis and machine learning algorithms to forecast potential outcomes, determine causation, assess risk exposure, and more. In the context of ERA, predictive AI typically involves the application of these trained models to interpret data gathered from sensors worn on the body; these trained models are used to determine whether workers are exposed to ergonomic risks based on the data received from the sensors on the workers [12]. However, it is crucial to acknowledge that while these sensor-based or marker-based predictive AI methods address the issues of low time efficiency, subjectivity, and inconsistency inherent in traditional approaches, they do not resolve the problem of intrusiveness [10]. On the other hand, computer vision (CV)-based predictive AI methods [5–9,13] can avoid introducing intrusiveness into ERA while overcoming the drawbacks of subjectivity, inconsistency, and time consumption. However, an unavoidable drawback of CV-based ERA methods is that occlusions in images or videos can significantly impact accuracy [13].

Unlike predictive AI, generative AI encompasses deep-learning models capable of processing raw data to "learn" and produce statistically probable outputs upon request. Fundamentally, these generative models encapsulate a simplified representation of their training data, which they utilize to generate new outputs that are similar yet distinct from the original data. The innovations brought about by transformers, particularly noticeable in large language models (LLM) tools such as ChatGPT, have ignited widespread interest, leading to investigations into their potential applications across various domains [14]. This surge in interest in LLMs using transformers suggests a range of promising research opportunities within the ERA field. In this context, existing ERA methods present two key limitations. First, predictive AI-based ERA methods currently struggle to accurately identify fundamental ergonomic issues and to provide corresponding solutions articulated in natural human language. Second, VLMs utilizing generative AI for ergonomic applications are insufficient, primarily due to a lack of specialized domain knowledge or relevant datasets.

Consequently, this study focuses on utilizing transformer-based LLMs to achieve high accuracy in assessing postural ergonomic risks based on image inputs, particularly within the construction industry, which is notably susceptible to WMSDs. Recognizing the necessity for image inputs in postural ERA, this research has chosen to utilize a vision transformer (ViT) for image processing. This approach enables distinct functionalities beyond those offered by predictive AI-based ERA methods. These functions include: 1) By generating human-like language descriptions as answers to visual questions, it informs users whether postural ergonomic risks are present in the input images. 2) By generating human-like language image captions, it provides descriptions of workers' actions in the input images and whether these actions might lead to postural ergonomic risks. It is noteworthy to emphasize the scarcity of publicly available datasets explicitly tailored for training or evaluating vision-language methodologies in generating descriptions of ergonomic risks that construction workers face. Therefore, this study has also created such a dataset. As a result, an interactive approach for postural ERA, ErgoChat, was developed to fulfill the two functions mentioned above. The ErgoChat is anchored in a vision-language model (VLM) with ViT architecture. Functionally, ErgoChat operates by mapping visual tokens derived from ViT's encoder into the feature space of an LLM. The underlying VLM is trained jointly on images and their associated textual descriptions, enabling it to comprehend visual content and translate this understanding into human-like language descriptions.

The contributions of this study include (1) an interactive method based on VLM that can recognize postural ergonomic risks and generate human-like language descriptions, (2) a dataset consisting of 1900 different image-text pairs designed explicitly for identifying postural ergonomic risks among construction workers, and (3) a comprehensive evaluation of the method was conducted using both quantitative and qualitative approaches, involving 9 distinct metrics and feedback from 50 ergonomic experts from around the world.

The remainder of this paper is structured as follows: Section 2 provides an overview of the relevant background on traditional and predictive AI-based ERA methods. The potential application of generative AI in ERA is also discussed. Section 3 introduces the proposed methodology. Section 4 details the analysis results and discussion. Lastly, Section 5 presents the concluding remarks and future work. The software, along with the dataset annotation, will be publicly accessible at https://github.com/xinmingliUofA/ErgoChat.

## 2 Related Work

Construction safety has long been an actively explored research field. Although injuries and deaths resulting from safety accidents are more catastrophic, the damage to workers' physical health caused by WMSDs due to ergonomic risks is a widespread issue that should not be overlooked [15]. As researchers and safety practitioners become increasingly aware of the prevalence of WMSDs caused by ergonomic risks in the workplace, more postural ERA tools have been proposed [16]. The widely recognized classification categorizes these postural ERA tools into self-reporting, manual observation, direct measurement, and more advanced technological methods [13]. Self-report, observation, and direct measurement approaches that do not incorporate AI are collectively referred to as traditional methods in the following text. ERA tools that employ predictive AI, such as those utilizing CV technology to analyze joint angles, are referred to as predictive AI-based methods. This section examines commonly used ERA tools in the construction sector.

2.1 Traditional ERA

**Self-report methods**. Self-report methods require researchers to collect ergonomic risk information from workers through surveys, interviews, or worker diaries [17]. The Standardized Nordic Questionnaire is an example of a self-report-based assessment method [18]. The Standardized Nordic Musculoskeletal Questionnaire was designed to identify ergonomic risks in the workplace by assessing the prevalence of WMSD. The Standardized Nordic Musculoskeletal Questionnaire was created for repeated screening and monitoring of WMSD and related ergonomic risk factors in the work environment, and it is not intended for clinical diagnosis [18]. A key advantage of the questionnaire is that it is standardized, allowing for the comparison of results across different studies. A more detailed version of the Standardized Nordic Musculoskeletal Questionnaire builds on the original one by adding questions about pain and the consequences of pain. The detailed version has been proven to be very reliable in identifying WMSD related to occupational and general population health [19]. Researchers and safety practitioners have also developed questionnaires tailored to particular industry sectors. Although many self-report methods have been proposed and even optimized, these self-report methods inevitably share common drawbacks. The drawbacks arise from the inconsistent perception, comprehension, or interpretation among those who respond to and process the questionnaires [19]. Other disadvantages include being time-consuming and prone to errors [20,21].

**Observational methods**. Observational methods evaluate the ergonomic risks associated with workers' postures using established tools according to the guidelines provided by these tools. Observational methods can be implemented through direct observation of workers or by analyzing video recordings of workers. The most common observation method-based assessment tools include Rapid Entire Body Assessment (REBA), Rapid Upper Limb Assessment (RULA), and the revised National Institute for Occupational Safety & Health (NIOSH) lifting equation [22,23]. REBA is an ergonomic assessment tool used to evaluate workers' risk of entire body WMSDs [24]. A key feature of REBA is its ability to rapidly assess the risk of WMSDs across the whole body, facilitating the improvement of worker tasks or interventions in improper movements. RULA is an ergonomic evaluation tool designed to assess workers' risk of upper body WMSDs [25]. Both RULA and REBA employ posture diagrams and scoring tables to identify potential sources of muscular fatigue. However, in contrast to REBA, a notable limitation of RULA is its exclusion of the lower limbs in its assessment, potentially overlooking WMSDs that could develop in the lower limbs. The Revised NIOSH Lifting Equation serves as an observational method for evaluating the physical demands associated with manual lifting tasks and assessing the potential risk of WMSD. The Revised NIOSH Lifting Equation output is a recommended weight limit calculated using six variables related to the task. The recommended weight limit defines the load an average worker can safely handle over a certain period [26]. Occupational health professionals use this tool to quantitatively analyze lifting risks and develop mitigation strategies. Table 1 provides a summary of some common observational methods. Like self-report methods, observational approaches also face challenges associated with human involvement in assessments, which can lead to inconsistencies, be time-consuming, and be prone to errors [13,20].

**Direct measurement methods**. Direct measurement methods rely on devices to monitor workers' postures, movements, level of surface activation, and forces exerted. These devices include but are not limited to, lumbar motion monitors, optical motion capture systems, non-optical motion capture systems, electronic goniometers, and heart rate monitors, among others [23]. Lumbar motion monitors quantify angles between body segments during movements by capturing the spine's position, velocity, and acceleration across three planes of motion, recording these parameters as functions of time [27]. Non-optical motion capture systems capture human movements without reflective markers or cameras; typically, these systems utilize inertial measurement units, magnetic field sensors, or a combination of mechanical linkages and potentiometers for motion capture [28]. In contrast, optical motion capture systems utilize reflective markers and cameras to record movements [28]. These systems employ infrared cameras that detect the light signals reflected from markers—typically small, spherical reflective balls—or from LED lights affixed to key points on the human body. It is important to highlight that optical motion capture systems, which use markers, offer greater freedom of movement and are less invasive than non-optical systems that employ sensors. However, the main drawbacks of optical systems are their dependency on high-quality ambient lighting conditions, and any obstructions between the markers and cameras can result in errors or loss of motion capture data [13]. Like Lumbar Motion Monitors, electronic goniometers also use sensors to calculate the angles between body segments [29]. Heart rate monitors use sensors to measure heartbeats, thereby calculating the amount of physical activity [30]. In summary, the disadvantage of sensor-based direct measurement methods is their intrusiveness, and particularly, the disadvantage of marker-based methods is that they impose stringent requirements on the workers' working environment.

Some researchers have employed simulation software or modeling approaches to assess ergonomic risk. However, these methods necessitate inputting workers' data from direct measurements into the software or models, thereby remaining susceptible to observer influence on the evaluation subjects. Several researchers have utilized such simulation tools to evaluate ergonomic risks among workers [31,32].

Traditional methods of postural ERA, such as self-reporting, observation, direct measurements, and simulation [4], are often constrained by their reliance on manual data collection. This process is time-consuming and usually leads to inconsistencies in data interpretation among ergonomists.

Table 1. Common ergonomic criteria proposed for traditional ERA.

| Observational Methods | Purpose | Body region considered | WMSD risk factors considered |
|---|---|---|---|
| REBA | Postural analysis for musculoskeletal risk | Trunk, neck, legs, knees, upper and lower arms, wrists | Awkward postures, load/ force, coupling, activity level |
| RULA | Assess exposure to risk of MSDs of the upper limb | Upper arms, lower arms, wrist, trunk, neck, legs | Repetition, awkward/ static postures, force, time worked without a break |
| NIOSH Lifting Equation | Control overexertion injuries caused by manual material handling and lifting | Low back | Lifting force, posture, repetition, duration |
| Washington State (WISHA) Lifting Calculator | Estimate the maximum safe lifting weight | Arms | Lifting force, duration, posture |
| OWAS | Assess exposure to risk of MSDs | Back, arms, and legs | weight of the load, posture |
| ULRA | Assess the upper limb, load resulting from repetitive tasks, | Upper limbs | Upper limb posture, force, duration, and repetitiveness. |
| OCRA | Quantifies the correlation between the daily number of actions performed by the upper limbs in repetitive tasks and the recommended number of actions. | Upper limbs | Repetitiveness, force, awkward posture and movements, and lack of recovery time. |
| NERPA | Assess exposure to risk of MSDs | Entire body | Awkward postures, load/ force, duration |
| LUBA | Assessment of postural loading of the upper body and limbs | Upper body and limbs | Posture, movement, external force, vibration, contact forces |

## 2.2 Predictive AI-based ERA

Predictive AI-based ERA uses various AI algorithms to learn from past data and analyze future data based on the knowledge acquired to perform risk assessments. Predictive AI-based ERA can be divided into three steps. The first step begins with collecting sufficient data to train the AI algorithms, which must be relevant to the desired prediction task. The second step involves using the data collected in the first step to train the AI algorithm, thereby continuously optimizing the model to accurately recognize patterns in future data. The third step involves deploying the trained AI model to predict outcomes based on its acquired knowledge from the historical data. Previous studies on Predictive AI-based ERA have commonly used AI techniques, including Support Vector Machines, Decision Trees, Long Short-Term Memory, Linear Regression, Artificial Neural Networks, Recurrent Neural Networks, and Convolutional Neural Networks, among others [12,33]. Significant potential for further exploration in deploying AI-based ERA methods within the construction industry remains. In contrast, the application of AI in ERA has been more extensively researched in general contexts. It is worth noting that these general methods can also be adapted for use in the construction industry through modification and improvement. This study only discusses construction industry-specific methods. Table 2 lists the recent studies on the use of predictive AI for ERA in the construction industry.

From Table 2, it can be observed that the majority of predictive AI-based ERA methods are implemented using sensors attached to the human body. Such methods share a common drawback - intrusiveness, which inevitably imposes psychological and physical burdens on workers. It is worth noting that the fundamental distinction between predictive AI-based and direct measurement methods lies in integrating AI for risk assessment. While both types utilize devices for data collection, predictive AI methods employ AI to automate the ERA process. In contrast, direct measurement methods rely on human evaluation for ERA. Approaches that use only video as the input for AI algorithms offer an alternative to sensor-based methods by eliminating the need to attach sensors or markers to workers. For instance, Fan et al. [13] have developed a technique using a CNN-based CV algorithm for the real-time tracking of workers' movements. This algorithm is trained using a previously collected construction workers' 3D dataset. Following this, the REBA method is applied to evaluate ergonomic risks from the predicted pose data obtained through the CV algorithm. The technique developed by Cai et al. [34] leverages a CV-based pose-tracking algorithm to track workers' movements. The next step is using an LSTM model to classify ergonomic poses, culminating in assessing these poses for ergonomic risks using the OWAS method. This methodology also demonstrates a non-intrusive way of conducting ergonomic assessments.

Table 2. Studies on AI-based ERA.

| Study | Data collection methods for training | AI algorithm | Ergonomic criteria | Task |
|---|---|---|---|---|
| Fan et al. [13] | Videos | CNN | REBA | Construction work activities |

| Antwi-Afari et al. [35] | Insole sensors | LSTM, Bi-LSTM, GRU | ISO 11226 (2000) | Body posture |
| --- | --- | --- | --- | --- |
| Cai et al. [34] | Videos | LSTM | OWAS | Body posture |
| Zhao et al. [36] | IMU sensors | CLSTM | OWAS | Construction work activities |
| Akanmu et al. [37] | IMU sensors | Reinforcement Learning, FL | PERA | Body posture |
| Antwi-Afari et al. [38] | Insole sensors | DT, RF, KNN, SVM, ANN | OSHA | Repetition |
| Umer et al. [39] | ECG sensor, Skin temperature sensor, Respiration sensor | KNNs, SVM, DAs, DTs, Ensemble classifiers | Borg-20 scale | Construction work activities |
| Zhao et al. [40] | IMU sensors | CLSTM | OWAS | Construction work activities |
| Zhang et al. [41] | IMU sensors | SVM | NS | Bricklaying activities |
| Yu et al. [42] | Pressure sensors, IMU sensors | NS | NS | Material handling, Rebar, Plastering |
| Nath et al. [43] | Body-mounted smartphone | SVM | OSHA | Material handling |

Abbreviations: CLSTM = Convolutional Long Short-Term Memory, DA = Discriminant Analysis, ECG = Electrocardiography, FL = Fuzzy Logic, GRU = Gated recurrent units, IMU = Inertial Measurement Unit, KNN = K-Nearest Neighbor, NS = Not Specified, OSHA = Occupational Safety and Health Administration, OWAS = Ovako Working posture Analyzing System, RF = Random Forest, PERA = Postural Ergonomic Risk Assessment.

2.3 Potential applications of generative AI in ERA

The fundamental difference between generative AI and predictive AI methods centers on their respective functionalities; generative AI is adept at generating new content, whereas predictive AI is primarily employed for classification or prediction tasks based on input data [44,45]. Generative AI distinguishes itself by its ability to produce original content, including texts, images, videos, and audio. This content creation capability allows for its innovative applications in ERA. When given specific prompts, generative AI has the potential to deliver straightforward yes or no responses to queries regarding the identification of ergonomic risks, elaborate on detected ergonomic risks with detailed descriptions, and even propose strategies to alleviate or prevent these risks. Nevertheless, predictive AI-driven ERA, which primarily concentrates on quantifying ergonomic risk levels through the detection of specific variables, has been limited in its ability to pinpoint the root ergonomic issues and propose corresponding remedies [46].

The application of generative AI in addressing ergonomic risks within the construction sector is underexplored, in contrast to its emerging utilization for safety risks. Searching "GPT" AND "ergonomics" on Scopus yields only two search results, and neither of these results is relevant to ergonomics. Searching "vision language" AND "ergonomics" on Scopus also yields only two search results, both authored by the same group of authors. Both of these articles propose a VLM capable of identifying ergonomic problems and their solutions [46,47]. The difference between them lies in the use of different VLM architectures.

Different from construction ergonomics, several generative AI methodologies have been proposed to address issues pertaining to non-ergonomic related construction safety. Smetana et al. [48] developed a methodology for text-based incident analysis utilizing the GPT-3.5 model from OpenAI. They applied GPT-3.5 and various natural language processing strategies to the OSHA severe injury reports database to pinpoint the leading causes of highway construction accidents. In a separate study, Uddin et al. [49] explored the potential of ChatGPT to bolster hazard recognition skills among students in construction-related disciplines, finding that ChatGPT significantly enhances their ability to identify hazards. Furthermore, Yoo et al. [50] proposed a transfer learning model and fine-tuned GPT to detect six distinct accident types. They also introduced a technique to improve the interpretability of the fine-tuned GPT model, involving saliency visualization on textual inputs. Nevertheless, while these approaches employ LLMs for text analysis and generation, they lack the capability to generate text responses from images or videos. In contrast to these generative AI methods that cannot process image inputs, Chen et al. [51] have proposed a visual construction safety query system that integrates with head-mounted virtual reality devices. This innovative system utilizes a VLM to produce image captions, facilitate VQA pertinent to safety, and conduct image-text retrieval tasks.

Notably, existing methods [46,47] employing generative AI to tackle ergonomic-related challenges exclusively rely on BLIP [52] as the foundational VLM. However, MiniGPT-v2 performs better than BLIP in VQA [53]. Moreover, the datasets utilized to train those two methods obtained textual descriptions of ergonomic risks by using ChatGPT-3.5. Utilizing ChatGPT for this purpose presents challenges stemming from its absence of domain-specific knowledge, reliance on input data quality, incapability to detect errors, lack of creative capacity, and limited comprehension of ethical considerations [54]. Additionally, both of these existing studies only employed the BLEU [55] metric to evaluate the performance of their proposed methods. Therefore, this study presents an interactive method to address the current scarcity of generative AI-based approaches for explainable IC to identify ergonomic issues and implement VQA in ergonomic risk recognition.

IC involves generating natural language descriptions of the visual content within an image, utilizing a visual perception system combined with a language model to produce coherent and syntactically accurate sentences [56]. In conventional IC, the task is treated as an image-to-sequence problem, where input pixels are encoded into one or more feature vectors during the visual encoding phase. This encoding serves as the input for the subsequent generation phase, where a language model generates a sequence of words. Recent advancements in transformers and self-attention mechanisms, coupled with methodologies akin to single-stream BERT, have enabled significant improvements in the design and performance of IC models [56].

## 3 Methodology

The proposed ErgoChat is an interactive chatbot capable of providing textual descriptions of the ergonomic risks faced by workers in an image input and identifying these ergonomic risks. The proposed ErgoChat can achieve VQA and IC regarding ergonomic risks for construction scenarios. Its architecture is adopted from MiniGPT-v2, comprising a visual backbone, a linear projection layer, and an LLM. Fig. 1 shows the architecture used by ErgoChat. The parts can be summarized as follows:

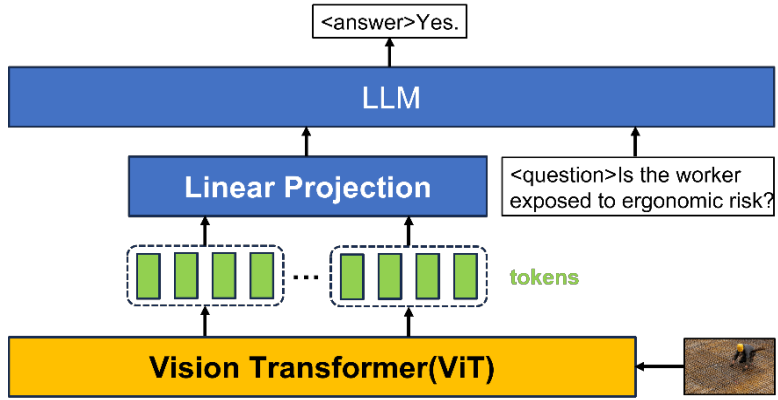

Fig. 1. The architecture used by ErgoChat

1) The visual backbone of ErgoChat utilizes the EVA ViT [57]. A ViT employs a self-attention mechanism for image processing. It is structured as a sequence of transformer modules, each containing two primary sub-layers: a self-attention layer and a feed-forward layer. The self-attention layer computes attention weights by evaluating the relationships among all pixels in the image, while the feed-forward layer performs nonlinear transformations on the outputs from the self-attention layer [57]. During the training of ErgoChat, this visual backbone remains frozen. An image resolution of 448x448 for training is employed. Additionally, interpolation techniques are utilized to handle position encoding, adapting to higher image resolutions.

2) The role of the linear projection layer in ErgoChat is to project visual tokens [57] from the frozen visual encoder into the feature space of the LLM [58]. The visual tokens and textual instructions/questions were provided/projected as inputs to the LLM for visual reasoning. Visual tokens represent the image patches into which an input image is divided, consisting of grids of fixed-size pixels. To mitigate the issue of decreased training and inference speed due to high-resolution images, ErgoChat employs a method similar to MiniGPTv2 [53] for handling visual tokens. These visual tokens are concatenated in groups of four adjacent tokens, and then these concatenated tokens are projected into the same feature space of the LLM. The feature space encodes diverse linguistic elements, including words, phrases, sentences, and extended text sequences, into numerical vectors that the model can process and analyze. As a result, this approach reduces the number of visual tokens to one-fourth of the original number, significantly enhancing training and inference speed when using high-resolution images.

3) ErgoChat adopts the open-source LLaMA2-7B [58] as the backbone for its LLM. The architecture of ErgoChat incorporates the LLM as an interface for inputs regarding vision-language tasks; ErgoChat utilizes language tokens from LLaMA2-7B to achieve vision-language VQA. ErgoChat can also perform functions other than VQA, such as IC/caption generation, grounded caption generation, and referring expression comprehension. However, this study focuses solely on exploring the performance of ErgoChat in VQA and IC related to ergonomic risk identification.

After establishing the architecture of ErgoChat, it becomes imperative to conduct thorough training and fine-tuning to enhance its accuracy in identifying ergonomic risks. This study utilized publicly available image-text pair datasets, although not specifically tailored for ergonomic risk, for the initial training of ErgoChat. The datasets used for pre-training and fine-tuning ErgoChat are listed in Table 4. Subsequently, to significantly bolster ErgoChat's proficiency in ergonomic risk identification, a dataset specifically designed for this task was proposed for fine-tuning the VLM. Following fine-tuning, various metrics were employed to quantitatively assess the performance of ErgoChat on the testing dataset proposed within this study. Additionally, questionnaires were distributed to experts to evaluate the performance of the IC qualitatively and quantitatively on the testing dataset. The research has obtained ethical approval (pro00111404) from the University of Alberta. Table 3 provides the training hardware and environment of the models.

Table 3. The training hardware and environment of the models.

| Hyperparameters/Environment | Value |
| --- | --- |
| GPU | 2 NVIDIA GA102GL RTX A6000 |
| CPU | Intel® Xeon(R) Gold 6242R |
| Operating system | Ubuntu 22.04 64-bit |
| GPU driver | Nvidia 555.42 with CUDA 12.5 |
| Programming language | Python 3.9 |

This study followed the five steps to ensure that ErgoChat possesses robust capability in identifying ergonomic risks and to test its accuracy. Fig. 2 illustrates the five steps.

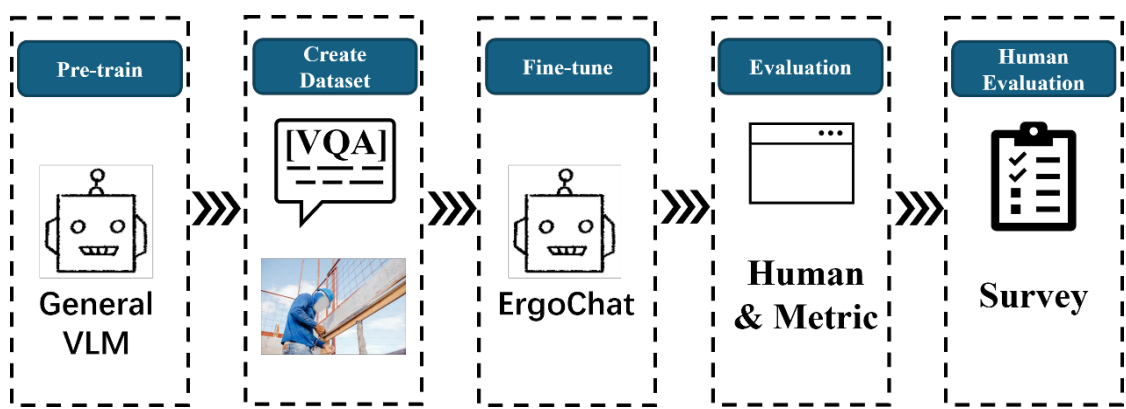

Fig. 2. The five steps of this study's workflow.

## 3.1 Pre-training ErgoChat with generic image-text pair datasets

ErgoChat was trained with publicly available generic image-text pair datasets not explicitly designed for ergonomic risk identification/knowledge, resulting in a large VLM with generic visual knowledge. The ErgoChat training using the generic dataset was divided into pre-train 1 to 3; the model underwent three pre-training sessions utilizing different combinations of those generic datasets. These datasets encompassed a broad spectrum of contents, including but not limited to humans, other animals, plants, architectural structures, vehicles, and landscapes. Pre-train 1 involves equipping ErgoChat with broad generic vision-language knowledge by leveraging a combination of numerous weakly-labeled image-text datasets and high-quality, fine-grained vision-language annotation datasets. Pre-train 2 involves refining the model using only fine-grained data for various tasks, such as VQA and IC. The rationale is to improve the model's performance in tasks such as VQA and IC. Therefore, the weakly-supervised dataset used previously was not utilized again in this training step. In pre-train 3, the model undergoes further fine-tuning with additional multimodal instruction and language datasets to enhance its capability to respond to diverse multimodal instructions and function effectively as a multimodal chatbot. The fine-tuning enables the model to gain more robust reasoning abilities while also addressing the issue where, after training on the GRIT-20M dataset (which contains very few grounded visual objects in its captions), the model could only visually recognize a limited range of objects [53]. Table 4 shows the pre-training and fine-tuning stages and datasets used in every stage. Fine-tuning utilizes the dataset proposed in this study and is discussed in the following sections.

Table 4. The pre-training and fine-tuning stages and datasets used in every stage.

| Dataset | Dataset type | Pre-train 1 | Pre-train 2 | Pre-train 3 | Fine-tune |
| --- | --- | --- | --- | --- | --- |
| GRIT-20M [59], LAION [60], CC3M [61], SBU [62] | Weakly-labeled | ✓ | ✗ | ✗ | ✗ |
| GRIT-20M [59] | Grounded caption | ✓ | ✗ | ✗ | ✗ |
| COCO caption [63], Text Captions [64] | Caption | ✓ | ✓ | ✓ | ✗ |
| RefCOCO [65], RefCOCO+ [66], RefCOCOg [67], Visual Genome [68] | REC | ✓ | ✓ | ✓ | ✗ |
| RefCOCO [65], RefCOCO+ [66], RefCOCOg [67] | REG | ✓ | ✓ | ✓ | ✗ |
| GQA [69], VQAv2 [70], OCR-VQA[71], OK-VQA[72], AOK-VQA [73] | VQA | ✓ | ✓ | ✓ | ✗ |
| LLaVA [74], Flickr30k [75], Multi-task conversation [53] | Multimodal instruction | ✗ | ✗ | ✓ | ✗ |
| Unnatural Instructions [76] | Language dataset | ✗ | ✗ | ✓ | ✗ |

| | | | | | | |
|---|---|---|---|---|---|---|
| Our fine-tuning dataset | Caption, VQA | ✘ | ✘ | ✘ | ✓ | |

## 3.2 Dataset curation for fine-tuning and testing

An image-text pair dataset was curated to fine-tune and test the VLM obtained in the last step. The partition for fine-tuning contained 1,700 distinct image-text pairs, and the partition for performance testing contained 200 image-text pairs. The images in both fine-tuning and testing partitions were obtained by searching 'construction works' in online resources with free-to-use licenses.

Descriptions of ergonomic risks in the text part of image-text pairs were based on using the REBA. The REBA employs a systematic approach to assess the upper and lower segments of the musculoskeletal system, identifying risks of WMSDs associated with a work task. The REBA classifies risk levels into five categories, spanning from negligible, low, medium, and high to very high risk [24]. The assessment of the presence of ergonomic risk in the textual descriptions associated with each image in the dataset relies on whether the REBA risk level of any worker depicted in the image equals or exceeds the medium risk threshold. If the REBA risk level meets or exceeds the medium risk threshold, the textual description will denote that the depicted worker is exposed to ergonomic risk; conversely, if the REBA risk level falls below the medium risk threshold, the textual description will indicate a lack of exposure to ergonomic risk. Alongside determining whether the workers depicted in the images are exposed to ergonomic risk, the text also describes the actions/tasks that lead to workers being exposed to ergonomic risk. A Python program was developed to convert these textual descriptions into a JavaScript Object Notation (JSON) file in MSCOCO format. The construction worker images and the JSON-formatted annotation file constitute this image-text pair dataset, which was utilized during the fine-tuning and testing of ErgoChat.

## 3.3 Fine-tuning ErgoChat with the specific dataset for ergonomic risk identification

The large VLM obtained from step 1 was fine-tuned with the fine-tuning dataset created in the second step. Using the ergonomic-specific dataset for construction workers proposed in this study, the fine-tuning serves to equip ErgoChat with knowledge of the ergonomic risks commonly faced by construction workers, thereby enhancing its performance in VQA and IC for ergonomic risks in the construction industry. This process yielded a large VLM with more ergonomic risk identification knowledge and capability. In this step, the ErgoChat pre-trained through the pre-train 1 to 3 underwent fine-tuning with the ergonomic risk-specific dataset. The dataset utilized in fine-tuning was precisely the fine-tuning partition created in step 2. In addition to the datasets used for pre-train 1 to 3, Table 4 shows the dataset used for the fine-tuning. After this fine-tuning stage, it was anticipated that ErgoChat would demonstrate notably improved performance in terms of ergonomic risk identification-related VQA and IC. Fig. 3 shows an example of the fine-tuning dataset, and the corresponding ground truth text by the ergonomic expert states: "The image shows a worker in a blue hoodie working on a fence. The worker's job duty requires a lot of physical activities. The worker is in a forward head posture and it is an ergonomic risk. The fence over the worker's head needs to be secured because it might fall on the worker and it is a safety hazard."

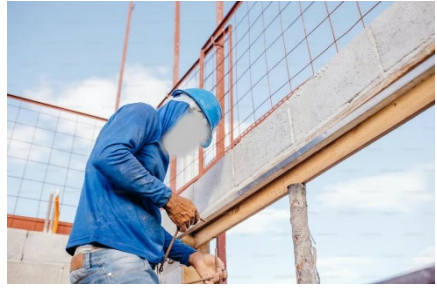

Fig. 3. A image from the fine-tuning partition [77]

3.4 Performance evaluation of the fine-tuned ErgoChat

The performance was measured from two perspectives: (1) accuracy in VQA related to ergonomic risk identification and (2) the ability to generate textual descriptions concerning ergonomic risk identification for workers.

3.4.1 Accuracy of VQA

The large VLM obtained after fine-tuning was tested with the testing dataset curated in step 2. The model obtained without fine-tuning (after pre-train 1 to 3) was also tested with the same testing partition so that the performance of ErgoChat before and after fine-tuning can be compared. In order to alleviate ambiguity encountered across various queries and tasks, the proposed dataset incorporates task identifier tokens designated for [caption] and [vqa]. In this dataset, [caption] is designated as the task identifier token for IC, while [vqa] serves as the task identifier token for VQA annotations. These identifiers are included at the beginning of the prompts to reduce ambiguity. Regarding VQA, the results before and after fine-tuning were compared with ground truth. For example, the VQA prompt for the test image in Fig. 4 was "[vqa] Is the worker exposed to postural ergonomic risks?" and both before and after fine-tuning, ErgoChat's responses were "yes." The same prompt "[vqa] Is the worker exposed to postural ergonomic risks?" was used for all the cases in the testing partition. Since the ground truth value for the VQA of Fig. 4 was "yes," the responses from ErgoChat both before and after fine-tuning were correct. Finally, the accuracy of VQA that the two models can achieve for the 200 test data points (percentage of correct answers out of the total) was calculated. Eq. 1 was used to calculate the percentage of correct risk identification using VQA.

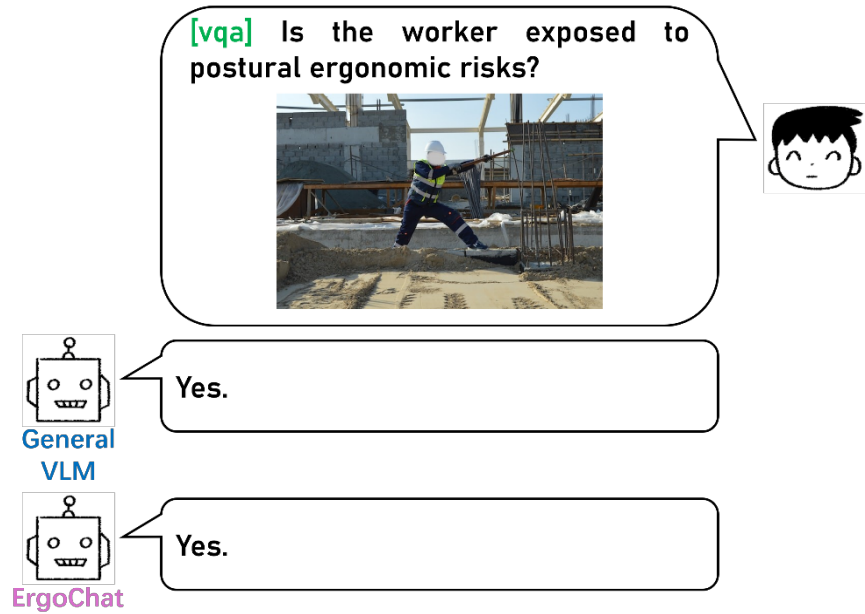

Fig. 4. The VQA question and answers from 2 models in an example

$$percentage\ of\ correct\ identication = \frac{number\ of\ correct\ identification}{200} \times 100\% \quad (1)$$

3.4.2 Quality of generated text descriptions

In terms of textual description of ergonomic risks generated by ErgoChat, the text generated before and after fine-tuning was compared with ground truth. The prompt used to create text descriptions was "[caption]Describe the workers and their postures in the image and tell me if they are exposed to ergonomic risks due to their postures?" The same prompt was used for all the cases in the testing partition. The performance of ErgoChat was measured by calculating the similarities between generated and ground truth descriptions. The perplexity [78] was used for VQA. A total of 9 metrics were used to calculate the similarities between the generated text and the ground truth text for every image in the testing dataset. The metrics were ROUGE_r, ROUGE_p, ROUGE_f [79], BLEU [55], NIST [80], cosine similarity [81], Euclidian distance [82], METEOR [83], and SPICE [84]. It is important to note that, except for perplexity, NIST, cosine similarity, and Euclidean distance, the range of other metrics is from 0 to 1. All these metrics except perplexity and Euclidean distance indicate better performance with higher scores. Table 5 provides more detailed information about these metrics. Perplexity was used to compare the VQA performance in this study, and the other 9 metrics were used to evaluate IC performance.

Table 5. The metrics used in this study to evaluate performance.

| Metric | Range | Description |
| --- | --- | --- |
| Perplexity [78] | 1 to ∞ | Assessment of how effectively the model has learned the distribution of the text on which it was trained. |

| ROUGE_r [79] | 0 to 1 | Measurement of the completeness of the information captured by a generated caption. |
| --- | --- | --- |
| ROUGE_p [79] | 0 to 1 | Indicator of the accuracy of the system-generated caption. |
| ROUGE_f [79] | 0 to 1 | A combination of ROUGE_r and ROUGE_p. |
| BLEU [55] | 0 to 1 | Measurement of how closely the machine-generated text resembles a set of high-quality reference texts. |
| NIST [80] | 0 to ∞ | Evaluation of machine translation that uses n-grams precision to calculate the similarity between a machine translation and a reference translation. |
| Cosine similarity [81] | -1 to 1 | A measurement of how similar the documents are, regardless of their size. |
| Euclidian distance [82] | 0 to ∞ | A measurement of the similarity between vectors, such as word embeddings or document vectors. |
| METEOR [83] | 0 to 1 | A measurement of the alignment between the generated text and the reference text, which assesses the quality of the generated text. |
| SPICE [84] | 0 to 1 | Evaluation of IC quality by analyzing the semantic similarity between reference and generated captions. |

The perplexity was used to test the probability that the generated textual descriptions for ergonomic risk identification were correct. If the generated description indicates that the worker is performing an ergonomically unsafe action, the corresponding score for identifying the ergonomic risk will be lower. A lower perplexity score indicates a higher probability that the description concludes the worker is exposed to ergonomic risk. A more detailed description of perplexity could be found in [78]. Perplexity is employed to estimate the likelihood of workers being exposed to ergonomic risks based on text descriptions generated by the two models compared in this study. When a text description's perplexity score indicates a higher probability of ergonomic risk and aligns with the ground truth, it implies that the VLM responsible for generating the description offers a more accurate image representation. ROUGE is a set of metrics designed explicitly for evaluating automatic summarization, which can also be applied to machine translation. The metrics compare a generated summary or translation against reference summaries or translations that are high-quality and produced by humans [78]. ROUGE_r was calculated by determining the ratio of unigrams present in both the ergonomics expert-generated description and the ErgoChat-generated description to the total number of unigrams in the ergonomics expert-generated description. ROUGE_p can be computed by taking the ratio of the number of unigrams in the ErgoChat-generated description that also appears in the ergonomics expert-generated description to the total number of unigrams in the ErgoChat-generated description. The ROUGE_f was calculated using ROUGE_r and ROUGE_p with Eq. 2. BLEU is a metric for comparing a generated translation to one or more reference translations. While initially developed for translation, it is also applicable for evaluating other tasks, such as paraphrasing and text summarization [55]. For further details on the other metrics used in this study, readers can refer to [80–84].

$$ROUGE\_f = \frac{2 \times (ROUGE\_r + ROUGE\_p)}{ROUGE\_r + ROUGE\_p} \qquad (2)$$

Due to disparities in the amount of ergonomic risk knowledge, the similarity between the textual descriptions produced by ErgoChat and the ground truth descriptions was expected to vary before and after fine-tuning. The nine previously mentioned metrics for IC were employed to evaluate the similarity between the generated textual descriptions and the ground truth descriptions for both general VLM (pre-trained 1 to 3) and fine-tuned (ErgoChat) models. Each of the 200 descriptions generated by the fine-tuned ErgoChat underwent evaluation for the nine metrics, yielding a total of 1600 computations. Similarly, the 200 descriptions generated by the pre-fine-tuned ErgoChat underwent the same evaluation. For each metric, the similarity scores of the general VLM were subtracted from those of the ErgoChat for each image, and the average of the differences for all 200 data points was determined by using Eq. 3. Subsequently, the average improvement in similarity across the nine metrics for the fine-tuned ErgoChat was computed. For each metric, the percentage of testing data that exhibited improvement with the fine-tuned ErgoChat was also determined using Eq. 4. Additionally, for the applicable metrics, the average improvement percentage by the fine-tuned ErgoChat across the 200 data points was determined using Eq. 5.

$$average\ of\ difference = \frac{\sum_{i=0}^{200}(f_{m,i} - u_{m,i})}{200} \qquad (3)$$

where $f_{m,i}$ means the fine-tuned metric score for the metric m and image i. Similarly, $u_{m,i}$ represents the un-fine-tuned (general VLM) metric score for the metric m and image i.

$$\%\ of\ data\ that\ indicated\ improvement = \frac{if\ (f_{m,i} - u_{m,i}) > 0, count + 1}{200} \qquad (4)$$

where $f_{m,i}$ means the fine-tuned metric score for the metric *m* and image *i*. Similarly, $u_{m,i}$ represents the un-fine-tuned (general VLM) metric score for the metric m and image i.

$$average\ improvement\ percentage = \frac{\sum_{i=0}^{200}(f_{m,i} - u_{m,i})}{200} \times \frac{1}{m_{max}} * 100\% \qquad (5)$$

where $f_{m,i}$ means the fine-tuned metric score for the metric m and image *i*. Similarly, $u_{m,i}$ means the un-fine-tuned (general VLM) metric score for the metric *m,* and image *i*. $m_{max}$ represents the maximum value/upper bound of the metric m.

3.5 Human evaluation of generated textual descriptions

In addition to the nine metrics, human evaluation was also employed to assess the improvement of fine-tuning on ErgoChat in terms of text description generation related to ergonomic risk identification. A questionnaire was devised to solicit assessments from individuals knowledgeable in ergonomics regarding the accuracy of descriptions generated by the post-fine-tuned ErgoChat versus those generated by the pre-fine-tuned ErgoChat for the testing dataset. A total of 50 participants/experts were recruited to complete the questionnaire. The questionnaire comprised 13 distinct versions, 8 of which included 15 different images from the testing dataset, while the remaining 5 versions contained 16 images. Together, these thirteen versions covered all 200 images from the testing dataset. Each image in the survey was paired with two questions. The first tasked participants with selecting the more accurate description of ergonomic risks faced by the construction worker depicted in the image from two provided text descriptions. The second question asked a participant's opinion on how much his/her choice is more accurate than the other one. These two text descriptions were generated by the pre-fine-tuned ErgoChat and the fine-tuned ErgoChat, respectively, and were identical to those in Section 3.4. The first question was formulated: "Which description is more accurate to describe the ergonomic risks in the image?" The choices for the first question were randomized for all the testing cases. Fig. 5 shows a question from the questionnaire. Additionally, the survey gathered demographic data on participants' highest level of education, the broad industry sector of their job, their experience level with ergonomic risk knowledge, and age.

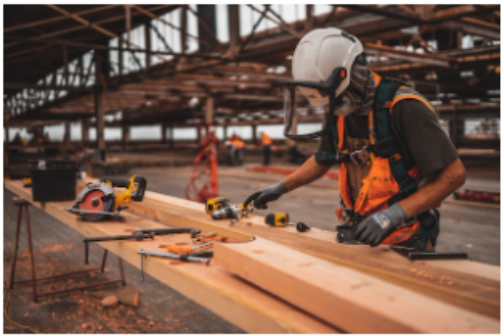

Fig. 5. A question from the questionnaire.

By analyzing responses to the first question, this study can gather evaluations from experts versed in ergonomics regarding the accuracy of descriptions produced by ErgoChat concerning ergonomic risks. Subsequently, these expert evaluations are compared with the assessments derived from the metrics outlined in Section 3.4 to ascertain the degree of concordance between the metrics and expert judgments. For example, suppose the majority of expert responses to the first question in the questionnaire suggest that descriptions generated by the fine-tuned

ErgoChat are more accurate, and simultaneously, results from the nine metrics indicate that these descriptions closely align with the ground truth. In that case, this provides further evidence supporting the superior performance of the fine-tuned ErgoChat in generating textual descriptions of ergonomic risk identification. In other words, this shows the methodology proposed in this study for identifying and describing ergonomic risks encountered by construction workers is effective. The process of interpreting the human evaluation results is shown in Fig. 6.

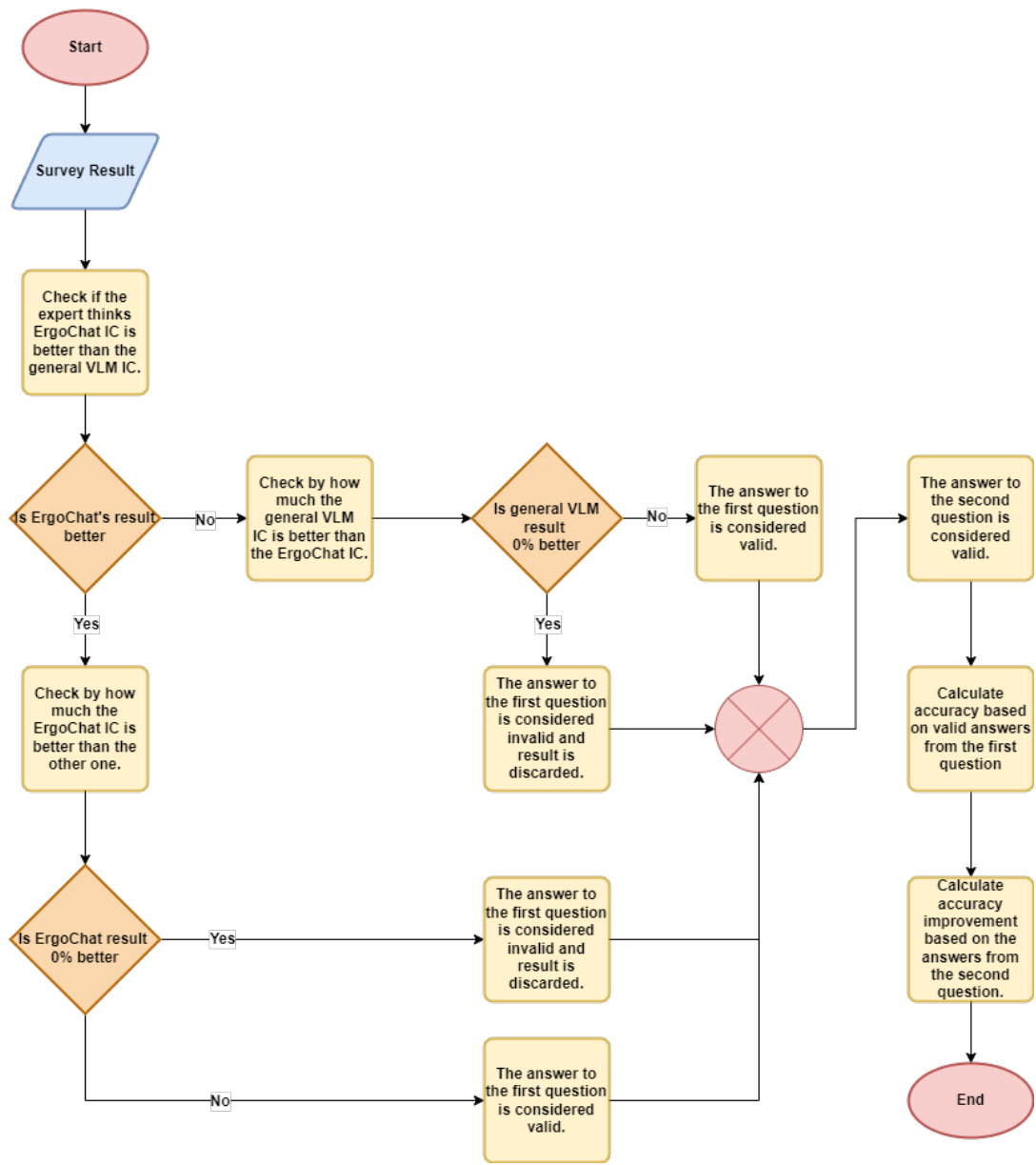

Fig. 6. The process of handling human evaluation data.

The second question requires participants to indicate the degree to which their chosen option in question one is more accurate than the other option they did not select. The phrasing of this question was, "By how much do you think your selected description is more accurate than the other? 0% (about the same) - 100% (absolutely more accurate)." The options for this question included 0%, 20%, 40%, 60%, 80%, and 100%. Analysis of the responses to this question can determine the extent to which experts perceive one option as more accurate. Concurrently, based on which ErgoChat model the most responses belong to, this study calculated the percentage by which one model is more accurate.

## 4 Results and Discussion

### 4.1 Performance of VQA

The testing dataset was utilized to assess the performance of the proposed ErgoChat in ergonomic risk identification in VQA. For the performance of ErgoChat and the general VLM models in ergonomic risk identification, this study calculated the accuracy of ergonomic risk identification by computing the results from the descriptions' perplexity scores and VQA. The results were obtained with the testing partition containing 200 data samples. Table 6 presents the accuracy of the general VLM and fine-tuned ErgoChat in identifying ergonomic risks using VQA and perplexity. The results were obtained with Eq. 1. Since ErgoChat generated text description's perplexity score indicates the probability of ergonomic risk, the perplexity scores from ErgoChat and the general VLM were calculated and compared to the ground truth of VQA. In the test dataset, 86% of the perplexity scores for descriptions generated by ErgoChat align with the ground truth, compared to only 63.5% for descriptions produced by the other VLM. The perplexity results indicate that the proportion of fine-tuned ErgoChat's outcomes matching the VQA ground truth is 22.5% higher than that of the other VLM. ErgoChat achieved an accuracy of 96.5% in responding to the prompt "[vqa] Is the worker exposed to postural ergonomic risks?" using only images from the test dataset. In comparison, the other VLM attained an accuracy of 95% on the same task.

Consequently, the proposed ErgoChat exhibits a slight improvement over the other VLM. It is worth noting that although the VLM used for comparison did not undergo fine-tuning, it demonstrated commendable performance in VQA for ergonomic risk identification. These findings suggest that the method, pre-trained on generic datasets, can determine whether workers face ergonomic risks in images. Nevertheless, its capacity to generate human-like descriptions of ergonomic risks and elucidate actions leading to such risks remains notably deficient.

Table 6. Percentage of correct risk identification in terms of VQA and perplexity

| | Correct identification | |
|---|---|---|
| **Metric** | **ErgoChat** | **General VLM** |
| perplexity | 86% | 63.5% |
| VQA | 96.5% | 95% |

### 4.2 Performance of IC of ergonomic risk-related descriptions

For the performance of fine-tuned ErgoChat and the general VLM in generating textual descriptions of ergonomic risks from images, this study utilized multiple metrics to calculate the similarity between the generated text and the descriptions provided by ergonomic experts. Table 7 shows the difference (improvement) in the average scores obtained by ErgoChat compared to the general VLM on the testing partition using 9 metrics and the percentage improvement represented by these differences. These two indicator types can be obtained using

Eq. 3 and Eq. 5. Additionally, Table 7 also includes the percentage of data on the entire testing partition for which improvement can be achieved through fine-tuning ErgoChat, and this can be obtained by using Eq. 4. Table 8 shows the average metric scores for fine-tuned ErgoChat and the general VLM. It is worth mentioning that among these metrics, only the Euclidean distance indicates lower similarity as the score increases.

Table 7. Metrics results of IC for the general VLM and ErgoChat.

| Metric | Average of difference (Eq. 3) | Percentage of data with improvement (Eq. 4) | Average improvement (Eq. 5) |
| --- | --- | --- | --- |
| ROUGE_r | 0.26 | 97.0% | 25.53% |
| ROUGE_p | 0.07 | 70.5% | 7.45% |
| ROUGE_f | 0.21 | 95.5% | 20.74% |
| BLEU | 0.14 | 95.5% | 14.44% |
| NIST | 1.45 | 97.0% | N/A |
| cos_similarity | 0.22 | 93.5% | 22.27% |
| Euc_distance | -0.42 | 65.0% | N/A |
| METEOR | 0.25 | 96.0% | 25.33% |
| SPICE | 0.16 | 93.5% | 16.30% |

Since cosine similarity ranges from -1 to 1, the values obtained from equations 3, 4, and 5 are not particularly informative. Moreover, a smaller Euclidean distance signifies higher accuracy in the generated text. Consequently, the data presented in Table 7 and Table 8 demonstrate that, apart from cosine similarity, other metrics indicate that ErgoChat is more accurate than the general VLM in delivering ergonomic risk-related information to construction workers. Results from 9 metrics employed to gauge the similarity between the generated descriptions of ergonomic risk identification and the ground truth affirm the superior accuracy of the proposed approach. Across the entire testing dataset, the average similarity scores obtained under the evaluation of the 9 metrics with the ground truth are higher for the proposed ErgoChat than the same architecture VLM. Specifically, the average ROUGE_r score of the proposed method surpasses that of the alternative method by 0.25531, indicating a 25.53% improvement because the ROUGE_r metric ranges from 0 to 1. Moreover, the proposed method demonstrates higher similarity scores for 97% of the data in the entire testing dataset than the alternative model. Corresponding results for other metrics are detailed in Table 7.

Table 8. Average metrics scores for IC across both models.

| Metric | Average of the general VLM | Average of ErgoChat |
| --- | --- | --- |
| ROUGE_r | 0.15 | 0.40 |
| ROUGE_p | 0.35 | 0.42 |
| ROUGE_f | 0.20 | 0.40 |
| BLEU | 0.01 | 0.16 |
| NIST | 0.17 | 1.61 |
| cos_similarity | 0.17 | 0.39 |
| Euc_distance | 8.97 | 8.55 |
| METEOR | 0.11 | 0.36 |
| SPICE | 0.10 | 0.26 |

For the evaluation of generated textual descriptions, in addition to the metric evaluation results in Table 7 and Table 8, the human evaluation results are summarized in Table 9. The process of handling the human evaluation results is illustrated in Fig. 6. The fine-tuned choice rate represents the average percentage of correct choice (the choice generated by the fine-tuned ErgoChat). The average accuracy improvement compared to the other choice represents the average improvement rated by the survey participants, and it represents the result for the following questions: "By how much do you think your selected description is more accurate than the other? 0% (about the same) - 100% (absolutely more accurate)." 54% of the questionnaire participants are Ph.D. degree holders or students; 40% are master's degree holders or students; 6% are undergraduate students or degree holders. Among the experts, 56% work in manufacturing, construction, or agriculture-related sectors, and 34% work in the public sector or education. 10% of the experts think they are experts in ergonomic knowledge; 16% think they are advanced; 28% think they are intermediate; 24% think they are novices; 22% believe they have fundamental awareness; None of them think they have no knowledge of ergonomics. 18% of the experts are 26 years old; 24% are 24, 25, or 29; 28, 32, or 34 years old take up 18%; the other ages do not comprise a significant portion of the experts. The ages of the experts range from 23 to 54.

Table 9. Results for human evaluation of ErgoChat generated ergonomic risk identification related text description.

| Metric | Fine-tuned choice rate | Average accuracy improvement compared to the other choice |
| --- | --- | --- |
| Value | 84.4% | 69.7% |

Through fine-tuning using the proposed fine-tuning partition, ErgoChat also demonstrated robust capabilities in ergonomic risk-related textual description generation when applied to real-world scenario-derived testing data. The proposed method exhibited improvements in ROUGE_r, ROUGE_p, ROUGE_f, BLEU, NIST, cosine similarity, Euclidian distance, METEOR, and SPICE for text generation compared to the VLM without fine-tuning. In the questionnaires, the experts were tasked with selecting the more accurate description from those generated by the two methods. Questionnaire responses from 50 experts with ergonomic knowledge indicated that 84.4% of the text descriptions generated by the fine-tuned ErgoChat were more accurate than those generated by the VLM without fine-tuning. Furthermore, the average accuracy improvement gain of descriptions generated by the proposed ErgoChat compared to those from the alternative VLM is 69.7%. Based on the results mentioned above, it is evident that the proposed method significantly surpasses the method with the same architecture but lacks fine-tuning in terms of ergonomic risk-related textual description generation. Fig. 7 illustrates an actual test case in the testing dataset. On the left side of the image are the questions from IC, along with the answers provided by ErgoChat, while on the right side are the questions from VQA, along with the answers provided by ErgoChat. Fig. 8 shows the IC and VQA generated by the other model for the same image.

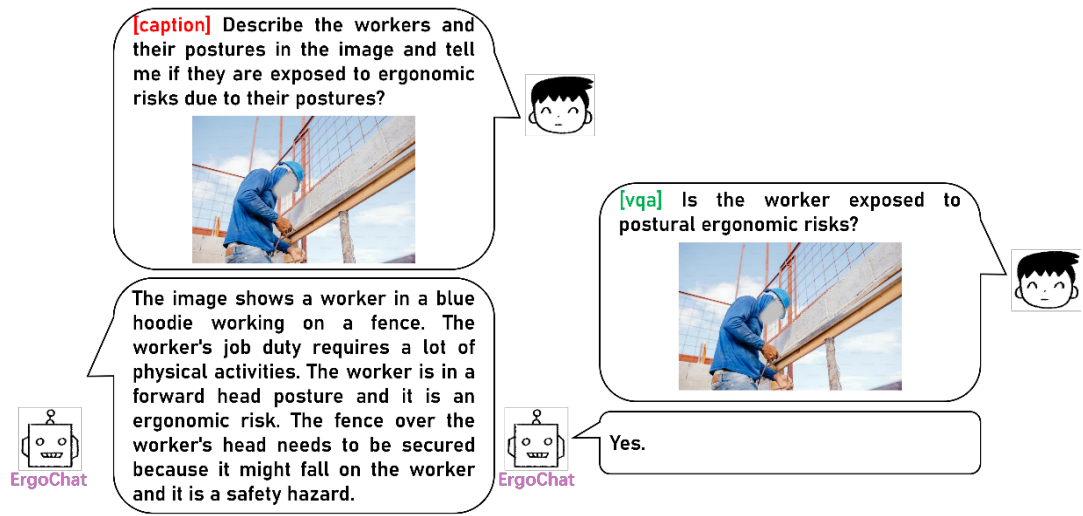

Fig. 7. An example of IC and VQA generated by ErgoChat from the testing partition.

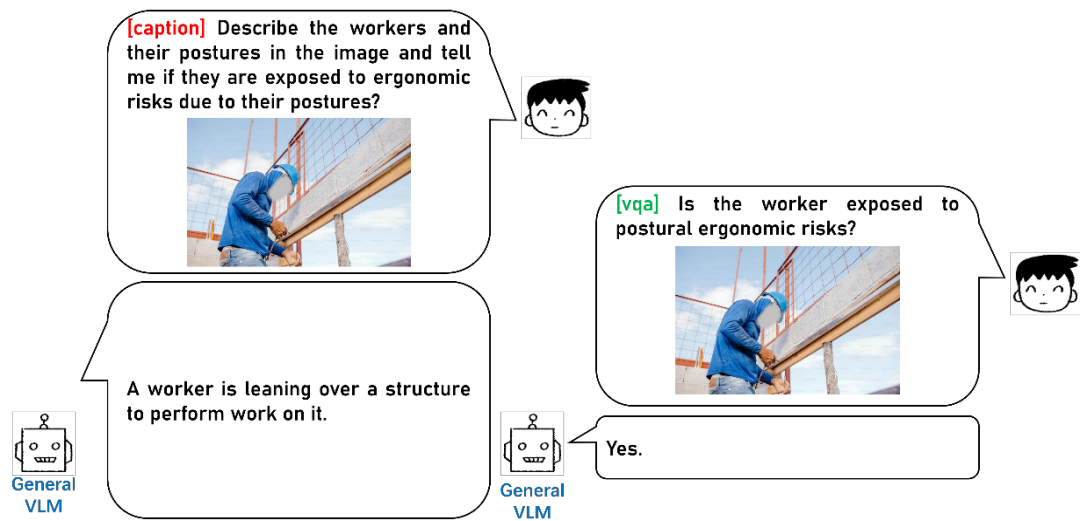

Fig. 8. IC and VQA generated by the other model for the same image.

In this study, the findings consistently showcase the remarkable performance of the proposed approach in generating human-like descriptions of ergonomic risks based on images and in VQA for identifying ergonomic risks. The proposed ErgoChat surpasses the VLM with the same vision-language architecture in both generating descriptions of ergonomic risks and VQA. Although the proposed method is highly effective, there is still room for improvement. Additionally, the text descriptions generated for the IC task sometimes incorrectly describe the actions of workers in the images, which can lead to failures in identifying ergonomic risks. These issues stem from ErgoChat's limited visual perception capabilities. They can be improved by training ErgoChat with a more extensive and more diverse set of image data.

ErgoChat was developed to aid construction safety practitioners in automating the detection of ergonomic risks associated with WMSD. This approach is designed to offer early alerts and address ergonomic risks and associated safety concerns encountered by construction workers. What distinguishes it from conventional AI algorithms used in identifying WMSD-related ergonomic risks is its capacity to produce descriptions of ergonomic risks akin to human language rather than solely determining whether workers are exposed to them. The significance of this lies in its ability to bridge the gaps left by traditional AI methods by providing human-like language descriptions. This capability allows ErgoChat to generate ergonomic injury reports on construction sites and support safety personnel in risk identification and safety inspections. Consequently, ErgoChat functions as an interactive tool for WMSD-related ergonomic hazard identification, facilitating ERA even for non-experts, and an automated tool for real-time hazard monitoring on construction sites.

## 5   Conclusion and Future Work

The primary objective in developing ErgoChat was to address the generation of textual descriptions concerning ergonomic risks linked to WMSD and VQA for ergonomic risk recognition in the construction sector. In these areas, traditional AI approaches have been limited in pinpointing the root ergonomic issues and proposing corresponding remedies. Built on the prevalent GPT architecture, similar methods still have significant room for improvements in the above functionalities, even after training solely on generic datasets. To explore the feasibility and potential for performance improvement of ErgoChat, this study introduced a dataset for fine-tuning and testing. To mitigate ambiguity across different queries and tasks, the dataset includes task identifier tokens for [caption] and [vqa]. The token [caption] is used as the task identifier for IC in the dataset, whereas the token [vqa] is utilized as the task identifier for VQA annotations.

One of our technical contributions is the proposed large VLM method for automatically or interactively identifying ergonomic risks related to WMSDs that construction workers face. Another technical contribution involves the creation of an image-text pair dataset designed explicitly for large VLM methods, incorporating task identifier tokens and addressing ergonomic risks faced by construction workers. The last technical contribution is a thorough evaluation of the method, incorporating both quantitative and qualitative approaches. This evaluation includes analysis across nine distinct metrics and insights gathered from 50 ergonomic experts. Unlike generic VLM methods, the proposed approach emphasizes the recognition and description of WMSD in construction workers in real-world scenarios.

The 9 metrics and human evaluation for IC all indicate that the proposed method outperforms the method of the same architecture in terms of ergonomic IC performance. Additionally, both the perplexity and VQA results demonstrate that the proposed method surpasses the method of the same architecture in ergonomic identification VQA performance. The evaluation experiments conducted in this study reasonably suggest that the practical application serves as a tool to enhance awareness of safety on construction sites and reduce the risk of WMSD among construction workers.

In the context of VQA evaluation, ErgoChat realizes a 1.5% increase in accuracy, achieving a total of 96.5%. Human evaluation results suggest that 84.4% of the text descriptions generated by ErgoChat are more accurate than those from the other model, with an average accuracy for these descriptions being 69.7% higher than that of the other model's outputs.

In terms of IC, the evaluation metric results reveal that while the average difference in cosine similarity between the two models is negligible, the remaining eight metrics indicate that

ErgoChat outperforms the other model in terms of IC. Specifically, six of these metrics demonstrate that ErgoChat achieves superior IC results for over 90% of the data in the test set. The other two metrics show improved IC for more than 65% of the test dataset. Human evaluation results demonstrated that 84.4% of the text descriptions produced by the fine-tuned ErgoChat model were more accurate than those generated by the unmodified VLM model.

As mentioned earlier, ErgoChat can offer early alerts of ergonomic risks and associated safety concerns encountered by construction workers. As an interactive tool, it can be used as an ergonomic risk query tool and to train construction site safety personnel who lack ergonomic risk knowledge. Since insurance companies and government occupational safety agencies both require injury reports, ErgoChat can be used to automatically generate ergonomic injury reports on construction sites. It can also assist safety personnel in fully automating risk identification and safety inspections. The ErgoChat proposed in this study is expected to play a role in identifying ergonomic risks for construction workers, thereby enhancing the well-being of workers and the safety of their work environment. Additionally, this study is anticipated to inspire further research on safety measures and VLMs to identify ergonomic risks.

Although this method exhibits significant accuracy, akin to applying visual language models in other domains, it also has limitations. These limitations pertain to challenges like language hallucinations and inadequate perceptual abilities. VLMs, built upon LLMs, inherit the LLM's limitations in language hallucinations primarily due to unreliable reasoning capabilities and a lack of understanding of non-existent hallucinations. The deficiency in perceptual abilities is attributed mainly to ErgoChat's restricted visual perception capacities. Additionally, we did not attempt different prompts. Prompt engineering is a separate research field, and it is not the focus of this study. Therefore, we did not explore the impact of different prompts on the results of IC and VQA. Future research endeavors could mitigate language hallucinations by leveraging high-quality image-text pairs containing ergonomic data and refined LLMs. To address the issue of limited perceptual capabilities, it might be necessary to integrate multiple layers within the projection layer to develop a more resilient visual perception model and employ more consistently coherent datasets.